# Recognition of *Pyralidae* Insects Using Intelligent Monitoring Autonomous Robot Vehicle in Natural Farm Scene

**Boyi Liu [1,2], Zhuhua Hu [1,\*], Yaochi Zhao [1], Yong Bai [1] and Yu Wang [1]**

[1] State Key Laboratory of Marine Resource Utilization in South China Sea, College of Information Science & Technology, Hainan University, Haikou 570228, China; liuboyilby@163.com (B.L.); yaochizi@163.com (Y.Z.); bai@hainu.edu.cn (Y.B.); yuwangfree@163.com (Y.W.)

[2] University of Chinese Academy of Science, Beijing 100049, China

\* Correspondence: eagler_hu@hainu.edu.cn; Tel.: +86-155-0090-9301

**Abstract.** The *Pyralidae* pests, such as corn borer and rice leaf roller, are main pests in economic crops. The timely detection and identification of *Pyralidae* pests is a critical task for agriculturists and farmers. However, the traditional identification of pests by humans is labor intensive and inefficient. To tackle the challenges, a pest monitoring autonomous robot vehicle and a method to recognize *Pyralidae* pests are presented in this paper. Firstly, the robot on autonomous vehicle collects images by performing camera sensing in natural farm scene. Secondly, the total probability image can be obtained by using inverse histogram mapping, and then the object contour of *Pyralidae* pests can be extracted quickly and accurately with the constrained Otsu method. Finally, by employing Hu moment and the perimeter and area characteristics, the correct contours of objects can be drawn, and the recognition results can be obtained by comparing them with the reference templates of *Pyralidae* pests. Additionally, the moving speed of the mechanical arms on the vehicle can be adjusted adaptively by interacting with the recognition algorithm. The experimental results demonstrate that the robot vehicle can automatically capture pest images, and can achieve 94.3% recognition accuracy in natural farm planting scene.

**Keywords.** Pest recognition, P*yralidae* pests, Agricultural robot, Image processing.

## Introduction

In agricultural production, the timely detection and identification of pests/diseases is a critical task for agriculturists and farmers to maintain the stability of grain output and reduce the environmental pollution caused by the usage of pesticides. *Pyralidae* insect is one of the most common pests in maize, sorghum and so on, and it does great harm to the quality and productivity of crops [1]. However, the traditional monitoring by humans requires a large amount of labor, and the detection is not timely due to human omissions. With the rapid development of artificial intelligence, computer vision-based pest detection has attracted attention from agriculturalists and computer scientists [2].

### *Related Work*

Some computer vision-based methods have been proposed to identify plant pests/diseases. Generally, they can be categorized as unsupervised and supervised methods.

In the category of unsupervised methods, the researchers need to analyze the specific features of pests/diseases, and design a feature extractor to separate the pests/diseases from background, then the additional features related to the segmented area are extracted to use for the classification of pests/diseases. For example, in [3], the authors applied multifractal analysis to segment action of whitefly images based on the local singularity and global image characteristics. In [4], convolutional Riemannian texture structure was explored to differentiate the environmental background textures and potential pest textures in pest camouflages in grains, then a differential entropic active contour model was developed to detect pests. In [5], to identify common greenhouse pests, such as whiteflies, aphid and thrips, the watershed algorithm was used to segment pests from the background images, and then color features of pests were subsequently extracted by Mahalanobis distance for identification of pest species. Besides single cue of feature channels, multi-feature fusion based method was also employed to identify pests/diseases. For example, in [6], the authors use color, shape and texture features and combined them with sparse representation to recognize pests.

In the past decade, as the increasing development of machine learning technology, more and more supervised models are designed to identify pests/ diseases. In traditional supervised learning, the features of pests/diseases are defined by human handcraft. In practical applications, the feature information is firstly computed, and then it is fed to suitable classifiers to recognize pests/diseases. For examples, Ali et al. applied DE color difference algorithm to separate the disease affected area, and the color histogram and textural features were also used to classify diseases [**7**]. Lu et al. used spectroscopy technology to detect anthracnose crown rot in strawberry [8], in which the spectral vegetation indices are used to train SDA (Stepwise Discriminant Analysis), FDA (Fisher discriminant analysis) and KNN(K-Nearest Neighbor) classifiers. Xie et al. employed KNN classifier to identify whether there is gray mold disease in tomato leaves from hyper-spectral images [9]. In [10], HOG(Histogram of gradient), Gabor and LBP (Local binary Pattern) features are calculated, and then Adaboost (Adaptive Boost) and SVM(Support Vector Machine) are used to recognize white-backed plant hoppers in paddy fields and identify their developmental stages. Similarly, in [11,12], the authors used histograms of oriented gradient features or region index as well as color index to design SVM structure, then SVM method is used for the classification of pests. In [13], features are extracted by using DCT (Discrete Cosine Transform) and then pests are classified with ANN (Artificial Neural Network). In [14, 15] the authors used the color feature to train a GMM (Gaussian Mixture Model), and with the trained model pest/diseases can be segmented accurately from background for subsequent recognition.

In the last five years, there are great innovations in the field of machine learning and DCNN (Deep Convolutional Neural Networks) [16, 17, 18]. They have dramatically advanced the object recognition to improve the recognition accuracy of pests/diseases in agricultures. For example, Sharada et al. applied deep learning technology to train a DCNN from 54,306 image of disease and healthy plant leaves collected under controlled conditions, and tested in their validation dataset with 99.35% recognition accuracy [19]. There are also pioneering work in training datasets [20, 21].

Regardless what supervised or unsupervised approach is, a large number of images with pests/disease are needed to construct and test a recognition model. However, the collection of pests/disease images is difficult and labor intensive. Researchers have been trying to reduce difficulty and labor force by pre-processing technology, such as focusing on or using stick trap [22, 3]. However, such designed algorithms are not suitable for the recognition of pests/disease in natural farm scene. Hence the acquisition technology of images with pest/disease is important for pest/disease recognition. We noted that there is little research to address this issue in present. In [23], the authors designed a pre-programmed autonomous pest control robot to sample *bemisia tabaci* adults, but the background of image is still not in a natural farm scene.

*Motivation and Main Contributions*

At present, most of the images of pests and diseases are taken manually, and some of them are selected for research. However, the effectiveness of the algorithms for continuous dynamic natural scenes has not been verified by many studies. In addition, most of the existing researches focus on simulation analysis for the detection of diseases and insect pests. The intelligent detection robots with autonomous control are rarely reported for the natural farm scene. Finally, the algorithms based on neural networks require a large number of training samples, which are not suitable for mobile embedded real-time monitoring devices. To overcome these shortcomings, we made the following contributions.

(1) We design a monitoring robot vehicle that can take images in natural farm scene. The recognition probability of diseases and insect pests is interacted in real time with the control the moving speed of the mechanical arms on the robot vehicle.

(2) We propose the detection and identification algorithm for *Pyralidae* pests. Considering that the color characteristics of adult *Pyralidae* pests are different from the background, we use color histogram mapping and constrained Otsu to accurately segment pests from complex scenes. Then, based on the segmentation results, we apply geometry invariant moments to recognize the *Pyralidae* pests in images. Our designed algorithm does not need large amount of data samples.

(3) We implement the algorithm on the embedded board of the smart vehicle, and verify the effectiveness of our proposed algorithm in the actual application scenarios.

The rest of this paper is organized as follows. Section 2 shows data acquisition equipment and its structure, and also describes of our proposed detection and recognition algorithm. In Section 3, we present the experimental results using the monitoring robot and the proposed recognition scheme. Finally, Section 4 concludes the paper.

**Materials and Methods**

*Acquisition Equipment of Pyralidae Insects Images*

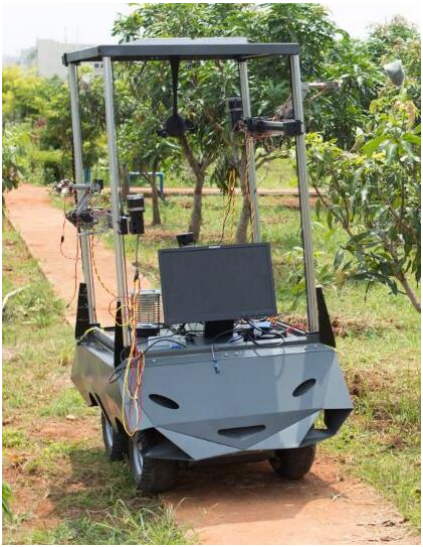 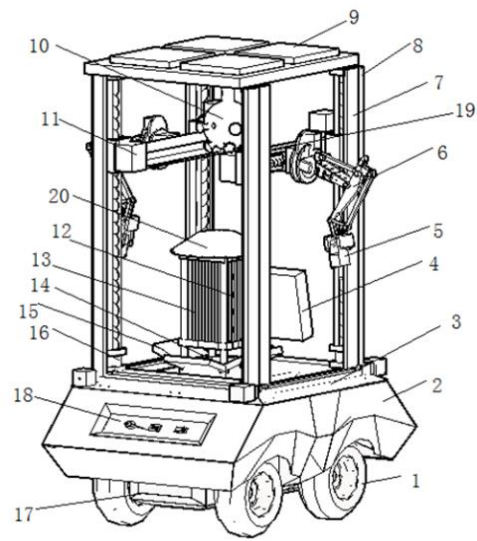

1.Deep grooved wheel, 2.Shell, 3.Guardrail, 4.Screen display, 5.Camera, 6. Mechanical arm, 7.Vertical thread screw, 8.Screw guardrail, 9.Solar panels, 10. Sensor integrator, 11. Horizontal screw motor, 12. Trap lamp, 13.The hardcore, 14. Cross bar, 15.Insects collecting board, 16.Vertical thread screw driven motor, 17. Chassis, 18. Vehicle control buttons, 19. Horizontal thread screw, 20. Trap top cover.

**Figure1. The intelligent autonomous robot vehicle**

The images used in our study are collected by an Automatic Detection and Identification System

for Pests and Diseases. The system was developed by our team and has been installed at the zone of technology application and demonstration of Hainan University in Hainan province, China. The system prototype and structure diagram is shown in Figure 1. The system can be divided into five major parts: the camera sensor (automatic focusing, resolution 1600×1200 and camera model KS2A01AF) and display unit; trap unit; the power transmission unit; the intelligent detection and recognition unit; and the hardware bearer unit. Table 1 shows the technical parameters of the robot vehicle.

**Table 1. The technical parameters of the robot vehicle.**

| Parameters | Value |
|---|---|
| Boundary dimension - (mm×mm×mm) | 1180×860×1800 |
| Running speed – (m/s) | 0.56 |

*Description of Proposed Recognition Scheme*

In this paper, the recognition scheme for *Pyralidae* insects, based on the method of inverse histogram mapping, is composed of five parts, including input module, reference image processing module, image segmentation module, contour extraction module, and object recognition module.

The input module firstly converts the experimental image into a matrix and initializes the parameters such as the binarization threshold of the probability image and contour recognition threshold. In the module of reference image processing, the reference image is transformed from RGB to HSV space and the histogram of the color layer (*H* layer) is computed. The image segmentation module is designed to compute the color histogram of the experimental image, normalize its histogram, obtain the probability image by mapping between histogram of test image and reference image, and then binarize the probability image.

Subsequently, in the contour extraction module, the contour of the binary image is extracted. The contours of the internal holes are removed by morphological methods according to the circumference and area features.

Finally, in the target recognition module, the algorithm recognizes the object contour by calculating the similarity between the contour obtained in the previous steps and the template contour. The object contour is confirmed when the similarity of the contour is larger than a threshold, and thus we can obtain the recognition results.

The pseudo-code for the proposed scheme is shown in Table 2. In Table 2, *S* denotes the template image, and *Mx* represents the image to be detected.

**Table 2. The pseudo-code description of proposed scheme.**

**Output**: Three vertices of triangular markings on the *Pyralidae* insects $(\alpha_1, \beta_1), (\alpha_2, \beta_2), (\alpha_3, \beta_3)$.

1: Initialize: (R, G, B)←*S, Mx*

2: Setting: The threshold of Hu moments; Reference contour image $Y_{image}$

3: V=max(R, G, B);

   S=(V-min(R,G,B))×255/V     if  V!=0,   0 otherwise

$$H = \begin{cases} (G-B) \times 60/S & if\ V=R \\ 180 + (B-R) \times 60/S & if\ V=G \\ 240 + (R-G) \times 60/S & if\ V=B \end{cases}$$

4: for i =0:1:255

   The color histogram of each image is obtained by statistics: H←$X_i$=$H_{pi}$/($H_m$×$H_n$);

```
         Normalize(H);
    end for
5: for i =0:1:m
    for j=0:1:n
        G_ij = Similarity(H of {Image blocks with the same size as M_x}, H of M_x)    /* Calculate the histogram similarity */
    end for
  end for
6: R=OSTU(G);         /* the image is binarized*/
7: C=findContours(R)        /*findContours() extracts the contours from binary images*/
8: real_match ← Based on Hu moment, calculating the similarity between C and the template contour
9: if   real_match > match:
        Triangle ← Approximate processing for triangle contour;
        Output vertex coordinates
   else：delete R
```

### *Probability Image Based on Color Histogram Inverse Projection and Multi-Template Matching*

The adults of the *Pyralidae* insects are yellowish brown. The back of the *Pyralidae* insects is yellowish brown, and the end of the abdomen is relatively thin and pointed. Usually, they have a pair of filamentous antennae, which are grayish brown. Meanwhile, its forewing is tan, with two brown wavy stripes, and there are two yellowish brown short patterns between the two lines. In addition, the hind wings of the *Pyralidae* insects are grayish brown; especially, female moths are similar in shape to male moths with lighter shades, yellowish veins, lightly brown texture and obese abdomen. From these characteristics, the color characteristics of adult *Pyralidae* insects are obvious, and it is very effective to recognize the *Pyralidae* insects by color characteristics. Color histograms are often used to describe color features, and are particularly useful for describing images that are difficult to segment automatically.

The inverse projection of the histogram is proposed by Michael J.Swain and Dana H. Ballard [24], which is a form of record that shows how the pixel or pixel block adapts to the histogram model allocation. It can be used to segment image or find interesting content in the image. The output of the algorithm is an image of the same size as the input image, where the value of the pixel represents the probability that it belongs to the object image. Therefore, it is possible to obtain a probability image by mapping the histogram in the object image by using the template image of the *Pyralidae* insects. Considering the *Pyralidae* insects' highlight color feature and the functional characteristics of histogram reflective algorithm, our proposed scheme applies image grayscale processing based on the reflection of the color histogram in the color feature extraction step. After the object image and the template image are converted into the HSV space and the color layer (i.e. the H component) is extracted, the image is grayed out by the method of histogram mapping. The gray image obtained in this way is a probability image that reflects the degree of similarity to the object color. Thus, it realizes the color distribution feature screening of the object image. The process of the algorithm is shown as follows:

(1) Convert the reference image into HSV space, extract the *H* spatial matrix, calculate statistical histogram and normalize it.

(2) Start from the first pixel (x, y) of the experimental image, cut a temporary image that is the

same size as the reference image, where (x, y) is the center pixel of the temporary image. Extract the $H$ space matrix, calculate its histogram and normalize it.

(3) Let $Similarity(H_1, H_2)$ denotes the similarity between the color histogram of the detected image $H_1$ and the color histogram of the reference image $H_2$. The degree of similarity indicates how the color characteristics of the pixels are in line with the probability of *Pyralidae* insects.

Calculate $Similarity(H_1, H_2)$ as described in Eq. (1) and Eq. (2).

$$H_k^{'} = H_k(i) - \frac{1}{N} \times \sum_{j}^{N} H_k(j) \tag{1}$$

$$Similarity(H_1, H_2) = \frac{\sum_{i}^{N} H_1^{'}(i) \times H_2^{'}(i)}{\sqrt{\sum_{i}^{N} H_1^{'2}(i) \times H_2^{'2}(i)}} \tag{2}$$

In the Eq. (1) and Eq. (2), $k \in \{1, 2\}$, $i = j \in \{1, 2, 3, \ldots, N\}$, $N$ is the number of intervals in the histogram, $H_k(i)$ is the value of the i-th interval in the k-th histogram.

In addition, the color and texture between different *Pyralidae* insects in natural scenes are distinct, and hence it is necessary to use a plurality of template images for histogram inverse projection processing. Otherwise, the use of a template cannot be adapted to a variety of different scenes. As shown in Table 3, three template images are given. The total probability image obtained is shown in Eq. (3), where $M$ represents the number of template images.

$$Similarity(H_1) = \sum_{m=1}^{M} Similarity(H_1, H_m) \tag{3}$$

After the images are converted to HSV space, histogram inverse mapping is conducted by employing three template images, and thus we can obtain the probability image respectively. Then, the three probability images are combined into one image, and logic and erosion operations are performed.

Table 3 shows the probability images obtained by histogram inverse mapping. The images in the first column are the template images. Except the template images, the rest images in Table 3 are experiment images or probability images. The first row is the original images containing *Pyralidae* insects, and the second to the 4th rows are the probability images obtained by mapping the test images' inverse histogram with the template image respectively. The last row is the total probability image

**Table 3. The original images and the obtained binary probability images after inverse histogrammapping.**

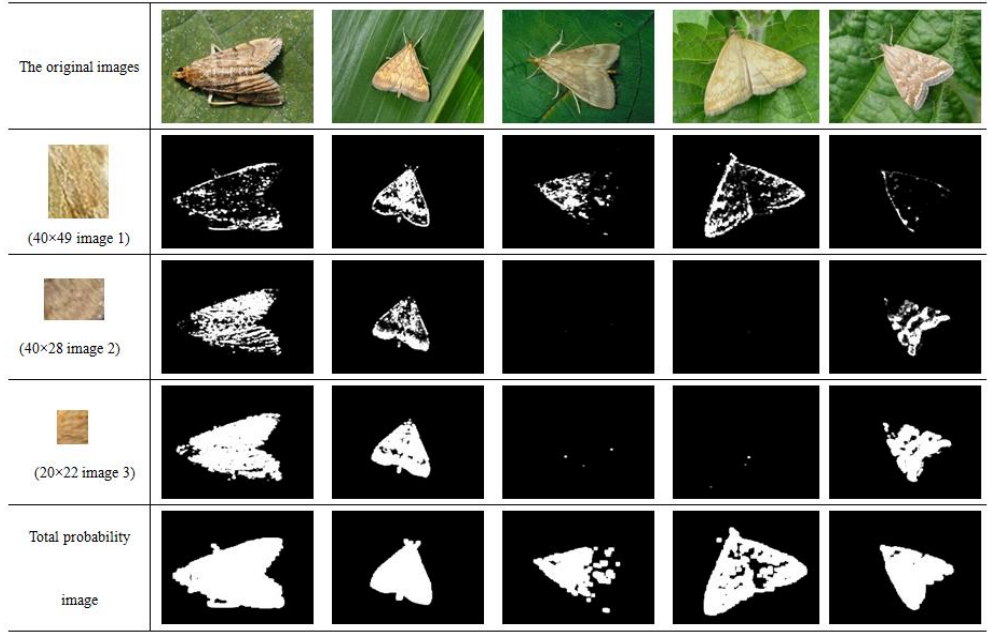

### Otsu Image Segmentation Based on Constrained Space

The Otsu algorithm is also known as maximum between-class variance method [25], which gives the optimal threshold to separate two classes in automatic image segmentation. For the image $G(x, y)$, the split threshold is set as $T$, $\omega_1$ is the proportion of foreground pixels, $\mu_1$ is the average grayscale of foreground image, $\omega_2$ is the proportion background pixels, $\mu_2$ is the average grayscale of background image, $\mu$ is the total average grayscale of background image, and $g$ is the maximum between-class variance. $p_{\min}$ and $p_{\max}$ are the minimum and maximum values of the pixel values in the image, respectively. Then, we can get

$$\mu = \mu_1 \times \omega_1 + \mu_2 \times \omega_2 \quad s.t. \quad \omega_1 + \omega_2 = 1 \tag{4}$$

$$g_{otsu} = \arg\max \left\{ \omega_1 \times (\mu - \mu_1)^2 + \omega_2 \times (\mu - \mu_2)^2 \right\} \tag{5}$$

Substitute the Eq. (4) into the Eq. (5), then the Otsu solution for threshold is expressed as

$$g_{otsu} = \arg\max \left\{ \omega_1 \times \omega_2 \times (\mu_1 - \mu_2)^2 \right\} \quad p_{\min} \leq T \leq p_{\max} \tag{6}$$

Finally, the maximum inter-class variance threshold of the image is obtained by iterative method. Inspired by [26], due to the difference in the similarity between the insects and the diversity of the natural scenes, the variance of the similarity of the background regions is small. In addition, the similarity of the *Pyralidae* insects is larger than that of the background. Therefore, the Otsu threshold is biased towards the background, which can lead smaller threshold compared with the actual optimal threshold.

$$g_{optimal} = \arg\max \left\{ \omega_1 \times \omega_2 \times (\mu_1 - \mu_2)^2 \right\} \quad g_{otsu} \leq T \leq p_{\max} \tag{7}$$

The constrained spatial Otsu segmentation method is used to obtain $g_{otsu}$, and then the threshold of maximizing the inter-classes variance is obtained in the constrained space (between $g_{otsu}$ and $p_{\max}$), as shown in Eq. (7), where $g_{otsu} = \left\lfloor \frac{1}{2}(\mu_1 + \mu_2) \right\rfloor$. $g_{optimal}$ indicates that the Otsu threshold is biased to a larger variance for the image with a large difference between the two variance values.

*Object Contour Recognition Based on Hu Moments*

The moment feature mainly characterizes the geometric characteristics of the image area, and it is known as the geometric moment. Because it has the invariant characteristics of the rotation, translation, scale and so on, it is also called the invariant moment. In image processing, geometric invariant moments can be used as an important feature to represent objects, which can be used to classify images. Among numerous geometric invariant moments, Hu moment is the most popular invariant moments [27].

Specifically, we assume that the gray distribution in the target $D$ region is $f(x, y)$. To describe the target, the gray distribution outside the region $D$ is considered to be 0, and then the geometric moment $m_{pq}$ and the regional moment $\mu_{pq}$ of the $p+q$ order are respectively expressed as follows:

$$m_{pq} = \iint_D x^p y^q f(x, y) dx dy \tag{8}$$

$$\mu_{pq} = \iint_D (x - \bar{x})^p (y - \bar{y})^p f(x, y) dx dy \tag{9}$$

According to the two above equations, a contour can be represented by $m_{pq}$ with $\mu_{pq}$. Based on the similarity between the experimental contour and the reference contour, the contour will be removed if the similarity is less than the threshold. The left contour is the contours of the *Pyralidae* insects. Finally, by using the function called approxPolyDP() in the OpenCV and other contour approximation processing function, the contour is approximated to a triangle and marked. Obviously, the marked contour is the desired result. *Remark*: *The approxPolyDP function can perform polygon fitting on contour points.*

*Recognition Algorithm interacted with Robot Control*

As can be seen from Figure 2, the mechanical arms on both sides carry cameras to obtain the video stream of diseases and insect pests in real time. Then, the video stream is transmitted to the analysis module in the robot vehicle, and the probability of the existence of *Pyralidae* insects in each frame is obtained, and the probability value is transmitted to the controller. If the probability exceeds a preset threshold, a warning message will be sent to the responsible experts. Meanwhile, the controller converts the received probability information into the speed of the motor in a linear relationship. There are four motors in the robot vehicle, namely two motors that control the up and down motion of the slide bar, and two motors that control the left and right motion of the mechanical arms.

Interaction of the robot operations with the pest recognition is one of the innovations of our proposal. According to the probability of similarity, the robot arms can adjust their speed. When the recognition similarity is greater than 0.9, the robot arms stop moving for the camera sensors to collect more images. Only when the similarity of five consecutive insect images is greater than 0.9, the robot issue an alarm and we can make the final decision on the presence of *Pyralidae* insects. When the similarity is between 0.7 and 0.9, the movement of mechanical arms slow down to account for the fact that maybe there is interference of other insects. By adaptively adjusting the moving speed of mechanical arms, the miss detection and false alarm can be significantly reduced.

In Figure 2, the robot vehicle can adjust the movement of the mechanical arms according to similarity probability. At the end of the arm, an embedded board with a camera sensor can change the angle of the shot. The mechanical arms also can move up and down or back and forth with the driving of the motor.

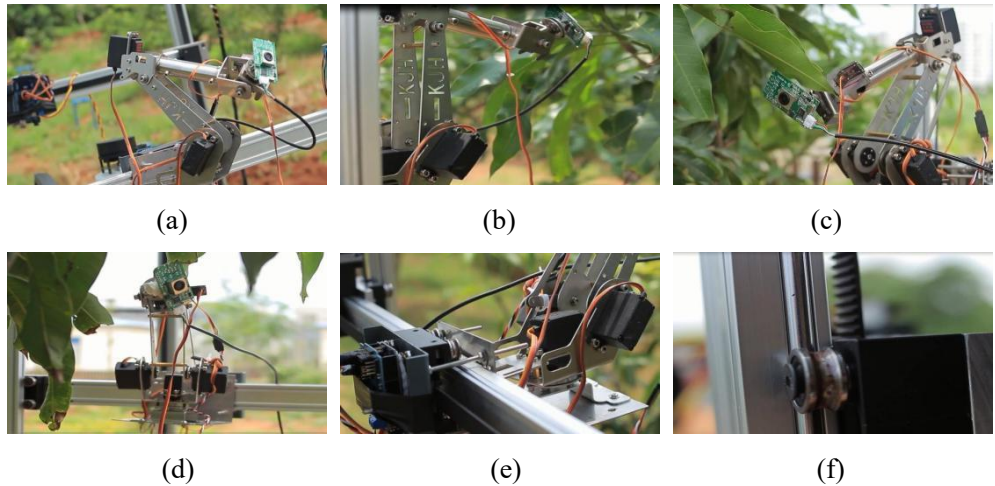

| (a) | (b) | (c) |
| (d) | (e) | (f) |

**Figure 2. Movement of the mechanical arm based on recognition similar probability.**
(a) - (d) Motion state of the mechanical arms; (e)-(f) The slide rails for the horizontal and vertical movement of the mechanical arms.

## Results and Discussions

The hardware platform of this scheme includes PC (Inter(R) Core(TM) i3-2500 CPU @3.30GHZ and 4.00GB RAM), embedded development board (NVIDIA Jetson TX1), embedded auxiliary control development board (2 Raspberry Pi B+ and 6 Arduino uno r3 expansion board), camera module (KS2A01AF), etc. The software used includes Window 7 operating system, Python 2.7, OpenCV 2.4.13 and embedded Linux operating system. The images used in the experiment are collected from cameras on the robot arms. We gather more than 200 images of the *Pyralidae* insects for experiments in the same conditions, and a few images of detection are shown in Table 4. The robot can perform a well-designed motion, capture clear images, and correctly identify the *Pyralidae* insects in the images.

### *Experiments and Analysis*

After the probability images are obtained, the contour extraction, matching, screening and recognition are performed. The triangles mark is used for recognition to characterize the *Pyralidae* insects shape. The recognition results are shown in Table 4.

**Table 4. The recognition results and the robot arm action.**

| | |
|---|---|
| (images of moths marked "decelerate" with similarity values 0.86, 0.72, 0.79, 0.79, 0.76, 0.81, 0.76, 0.86, 0.75, 0.89, 0.71, 0.79) | When the experimental results of the left images are presented, the robot arm will slow down. Thus, the number of samples in the region can be increased to improve accuracy. |
| (images of moths marked "stop" with similarity values 0.97, 0.91, 1, 1, 0.90, 0.91, 0.96, 0.91) | When the experimental results of the left images are presented, the robot arm will stop. More images will be collected and the alarm will go off. |
| (images marked "hold on" with values 0.69, 0.67) | The experimental results of the left images are presented, the robot arm will hold on. |

As can be seen from Table 4, our proposed scheme can identify the object when the images contain *Pyralidae* insects. The number illustrated on the pictures indicates the similarity value (e.g., 0.86). When we use the triangle for identification, better results are achieved. According to different recognition results, the speed of the robot arms can be adjusted adaptively. The processing time is about one second on each image.

In order to verify the advantages of the proposed scheme in the identification of *Pyralidae* insects, we compare with the three existing methods, namely, Support Vector Machine (SVM) method [12], the multi-structural element-based crop pest identification method proposed in [28] and the general histogram reverse mapping method. The experimental results based on SVM are shown in Table 5.

**Table 5. Recognition results and time costs are based on the SVM method.**

| | |
|---|---|
| The results of correct identification are based on the SVM method. | (12 images of moths labeled "classification result:positive" with correct rates: 0.78947, 0.47368, 0.63158, 0.47368, 0.52632, 0.63158, 0.52632, 0.47368, 0.47368, 0.84211, 0.57895, 0.57895) |
| Time cost (/s) | 3.10, 2.10, 1.85, 3.20, 3.12, 3.20, 2.80, 3.10, 2.80, 3.10, 2.87, 3.21 |

| The results of incorrect identification are based on the SVM method. | 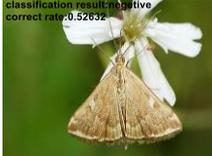 |
|---|---|
| Time cost (/s) | 3.22, 2.95, 2.99, 3.01, 3.24, 1.96, 2.75 |

From the experimental results, the recognition accuracy of the model trained by SVM method is about 65%, which is lower than the accuracy of the method proposed in this paper. Moreover, the accuracy of hyperplane obtained by training is about 0.5, which leads to the unstable recognition. The average time cost of the SVM method spent on an image is about 2.87s.

The experimental results are shown in Table 6. The recognition scheme of *Pyralidae* insects proposed in this paper has higher recognition accuracy and lower false alarm rate. Besides, it is not necessary to carry out a large amount of data analysis, which ensures that the average time-consuming is not significantly increased. In Table 6, the recognition accuracy and the false positive value are calculated as follows:

$$\beta = \frac{\sum_{i=1}^{n} r_{ij}}{n} (r_{ij} = 0 \ or \ r_{ij} = 1) \tag{10}$$

$$\delta = \frac{x - \sum_{i=1, j=1}^{n,m} r_{ij}}{x} (r_{ij} = 0 \ or \ r_{ij} = 1) \tag{11}$$

In (10) and (11), $\beta$ represents the recognition rate and $\delta$ represents the false positive rate. $r_{ij}$ is the j-th contour of the *i*-th *Pyralidae* insects( If exist, then 1, else 0). $n$ represents the number of real *Pyralidae* insectss in the image, $x$ represents the total number of contours marked by the algorithm, and m represents the total number of contours marked by the algorithm for the *i*-th *Pyralidae* insects in the image. Thus, the recognition accuracy reflects the ability of the algorithm to identify *Pyralidae* insects. The false alarm rate reflects the proportion of the error contours in all marked contours. Especially, the sum of these two probabilities is not necessarily equal to 1.

Our scheme and other three existing algorithms are used to test more than 200 images containing the *Pyralidae* insects, respectively. Then, we conducted a statistical analysis for the average time consumption, the recognition accuracy, the false alarm rate. The results of the statistics are shown in Table 6.

**Table 6. Comparison results of different schemes.**

| Methods | Recognition accuracy rate (%) | False alarm rate (%) | Average time-consuming (/s) |
|---|---|---|---|
| The proposed method | 94.3 | 6.5 | 1.12 |
| Color histogram backprojection [29] | 65.2 | 60.8 | 1.01 |
| Multi-structural element method [28] | 78.8 | 16.9 | 1.10 |
| SVM method [12] | 65 | 49.3 | 2.87 |

*Discussions*

Currently, recognition method based on deep learning has a rapid development. Unfortunately, the capture and establishment of pest images of borers are very difficult. The available images are far less than the requirement for training a deep learning network model. In addition, the robot can adaptively adjust the sampling frequency to detect with our proposed scheme.

*Future Work*

Firstly, the average processing time of each frame is slightly longer than 1 second, which will lead to a lag in the response of the robot to the observation results. In the future work, we will optimize the algorithm and improve the processing capacity of hardware to reduce the time overhead of video frame processing. Secondly, in the natural farm scene, due to the non-uniformity of the illumination, the local color of the images can be too bright or too dark, and it is difficult to determine a relatively stable threshold range, which affects the identification accuracy of pests and diseases. Therefore, the detection of pests and diseases under non-uniform illumination needs further study.

**Conclusions**

*Pyralidae* pests have great harmful impacts on the quality and productivity of crops. In order to achieve the accurate detection and identification of *Pyralidae* pests, we designed an intelligent robot on an autonomous vehicle to acquire images in practical farm scene, and we also presented a *Pyralidae* pests' recognition algorithm to be used in the robot. Specifically, by employing the color and shape characteristics of *Pyralidae* pests, we proposed a segmentation algorithm by inverse histogram mapping and constrained Otsu to segment pests. Then, we designed a recognition approach based on Hu invariant moment. Finally, compared with existing methods, the accuracy of our recognition scheme can reach to nearly 94.3% with acceptable time complexity. Another advantage of our proposed scheme is that it does not require large number of images of *Pyralidae* pests to be collected.

**Acknowledgements**

This research was supported by Key R & D Project of Hainan Province, China (Grant No. ZDYF2018015), Hainan Province Natural Science Foundation, China (Grant No. 617033) and Oriented Project of State Key Laboratory of Marine Resource Utilization in South China Sea, China (Grant No. DX2017012).